\documentclass{new_tlp}
\usepackage{mathptmx}
\usepackage{amsfonts}
\usepackage{amsmath}
\usepackage{amssymb}%\hyphenation{op-tical net-works semi-conduc-tor}
\usepackage{graphicx}
\usepackage{caption}
\usepackage{subcaption}
\usepackage{float}
\usepackage{url}
\usepackage{hyperref}
\usepackage{ntheorem}

\theoremstyle{nonumberbreak}
\newtheorem{example}{Example}

\newcommand{\cphydra}{CPHydra}

\newcommand{\satzilla}{SATzilla}
\newcommand{\sunny}{\text{SUNNY}}
\newcommand{\solver}{\texttt{sunny-csp}}
\newcommand{\tool}{\texttt{mzn2feat}}

\usepackage{listings}
\makeatletter
\def\url@leostyle{%
  \@ifundefined{selectfont}{\def\UrlFont{\sf}}{\def\UrlFont{\small\ttfamily}}}
\makeatother
\begin{document}
\title{SUNNY: a Lazy Portfolio Approach\\for Constraint Solving}

\author[Amadini and Gabbrielli and Mauro]
         {ROBERTO AMADINI \ and \ MAURIZIO GABBRIELLI \ and \ JACOPO MAURO \\
          Department of Computer Science and Engineering/Lab. Focus INRIA, University of Bologna, Italy.}
          
\maketitle

\begin{abstract}
\textbf{*** To appear in Theory and Practice of Logic Programming (TPLP) ***}

Within the context of constraint solving, a portfolio approach allows one to exploit 
the synergy between different solvers in order to create a globally better solver.
In this paper we present \sunny: a simple and flexible algorithm that 
takes advantage of a portfolio of constraint solvers in order to compute --- without learning an 
explicit model --- a schedule of them for solving a given Constraint Satisfaction Problem (CSP).
Motivated by the performance reached by \sunny\ vs.~different 
simulations of other state of the art approaches, we developed \solver, an 
effective portfolio solver that exploits the underlying \sunny\ algorithm in order to solve a given CSP. 
Empirical tests conducted on exhaustive benchmarks of MiniZinc models show that the 
actual performance of \solver\ conforms to the predictions. This is encouraging both for 
improving the power of CSP portfolio solvers and for trying to export 
them to fields such as  Answer Set Programming and Constraint Logic Programming. 

\end{abstract}

  \begin{keywords}
    Algorithms Portfolio, Artificial Intelligence, Constraint Satisfaction, Machine Learning.
  \end{keywords}

\section{Introduction}
\label{sec:introduction}
Constraint Programming (CP) is a declarative paradigm that enables expressing
relations between different entities in the form of constraints that must be satisfied. 
One of the main goals of CP is to model and solve \emph{Constraint Satisfaction Problems} (CSP) 
\cite{DBLP:journals/ai/Mackworth77} as well as problems like 
the well-known Boolean satisfiability problem (SAT), Quantified Boolean Formula (QBF), 
Satisfiability Modulo Theories (SMT), and Answer-Set Programming (ASP).
One of the more recent trends of CP --- especially in the SAT field --- is trying to 
solve a given problem by using a \emph{portfolio} 
approach~\cite{rice_algorithm_selection,DBLP:journals/ai/GomesS01}.
% \cite{DBLP:journals/ai/GomesS01} - This reference is misleading, because the type 
% of (parallel) portfolio approach described by Gomes & Selman has not had as much 
% traction on SAT as algorithm selection based on portfolios. In particular, 
% their concept of portfolio is not what you describe later in the same paragraph.

A portfolio approach is a general methodology that exploits a number of 
different algorithms in order to get an overall better algorithm. A portfolio of 
CP solvers can therefore be seen as a particular solver, dubbed a \textit{portfolio solver}, 
that exploits a collection of $m > 1$ different constituent solvers $s_1, \dots, s_m$ 
in order to obtain a globally better CP solver.
When a new unseen instance $i$ comes, the portfolio solver tries to predict which are the 
best constituent solvers $s_1, \dots, s_k$ (with $1 \leq k \leq m$) for solving $i$ and then runs such solver(s) 
on $i$. This solver selection process is clearly a fundamental part 
for the success of the approach and it is usually performed by exploiting \textit{Machine 
Learning} (ML) techniques.
Exploiting the fact that different solvers are better at solving different problems, 
portfolios have proved to be particularly effective. For example, the overall winners of 
international solving competitions like the \citeN{sat_competition} and \citeN{CSC09} are 
often 
portfolio solvers.

Surprisingly, despite their proven effectiveness, portfolio solvers are rarely 
used in practice and usually restricted to the SAT field. 
Outside the SAT world there are indeed only few applications of portfolios: 
for example, \citeN{asp_portfolio} for ASP, \citeN{cphydra} for CSP, and 
\citeN{prediction_state_art} for optimization problems like the Traveling 
Salesman Problem. On the other hand, if we exclude competitive scenarios such 
as the SAT Competitions, even the SAT portfolio solvers are actually underutilized.
As pointed out also by \citeN{snappy}, we think that one of the main reasons for this 
incongruity lies in the fact that state of the art portfolio solvers usually 
require a complex off-line training phase and they are not suitably structured 
to incrementally exploit new incoming information.
For instance, a state of the art SAT solver like 
SATzilla~\cite{satzilla} requires a model built by exploiting a 
weighted Random Forest machine learning approach while 
3S~\cite{3s}, ISAC~\cite{DBLP:conf/cpaior/MalitskyS12}, 
and CHSC~\cite{cshc} during their off-line phase cluster 
the instances of a training set or solve non-trivial combinatorial problems 
for computing an off-line schedule of the solvers.
In order to use these solvers a suitable set of training instances must be found 
and, especially, the off-line phase has to be run on all the training samples. 
Moreover, new information (e.g., incoming problems, solvers, features) may 
imply a re-running of the whole off-line process. 

In the literature some approaches avoiding a heavy off-line training phase have been 
studied \cite{Wilson00,pulina07,cphydra,NikolicMJ09,milano04,pulina10,snappy}. As pointed 
out in 
the comprehensive survey by \citeN{selection_survey}, these 
approaches are also referred as \textit{lazy}.
% \footnote{More precisely, ``\emph{Lazy approaches 
% do not learn an explicit model, but use the set of training examples as a case base. 
% For new problems, the closest problem or the set of $n$ closest problems in the case
% base is determined and decisions made accordingly.}''~\cite{survey}}
Unfortunately, to the best of our knowledge, among these lazy approaches only CPHydra 
\cite{cphydra} 
was successful enough to win a solver competition in 2008. Recently, however, 
\citeN{cpaior} showed that some non-lazy approaches 
derived from SATzilla \cite{satzilla} and 3S \cite{3s} have better performance than CPHydra on 
CSPs.

Our goal is, on one hand, bridging the gap between SAT and CSP portfolio solvers, 
and, on the other hand, encouraging the dissemination and the utilization in 
practice of portfolio solvers even in other growing fields such as, for example, the 
Answer-Set Programming and Constraint Logic Programming paradigms 
where a few portfolio approaches have been studied (see for example \citeN{asp_portfolio}).
As a first step in this direction, in this paper we present
\sunny\ and \solver. 

\sunny\ is a lazy algorithm portfolio which exploits instances similarity to 
guess the best solver(s) to use. 
%\sunny\ is the acronym of its main components: SUb-portfolios, Nearest Neighbors algorithm and lazY approach.
For a given instance $i$, \sunny\ uses a \textit{$k$-Nearest Neighbors} ($k$-NN) algorithm 
to select from a training set of known instances the subset $N(i, k)$ of 
the $k$ instances closer to $i$. Then, it creates a schedule of solvers 
considering the smallest \textit{sub-portfolio} able to solve the maximum number of 
instances in the neighborhood $N(i, k)$. The time allocated to each solver of the 
sub-portfolio is proportional to the number of instances it solves in $N(i, k)$. We performed a preliminary evaluation of the performance of \sunny\ by  
exploiting a large benchmark of CSPs from each of which we extracted an 
exhaustive set of features (e.g., number of constraints, number of variables, domain sizes) used to 
estimate the instances similarity.  
Following the methodology of \citeN{cpaior}, we measured 
the performance of \sunny\ by comparing its expected behavior against the 
simulations of state of the art portfolio solvers, namely, CPHydra, 3S, and 
SATzilla.
%Designed initially for reasons of flexibility and usability, we realized that 

The performance of \sunny\ were promising: it overcame 
all the above mentioned approaches. We therefore developed 
\solver, a new CSP portfolio solver built on the top of the \sunny\ algorithm. 
Using the very same benchmarks and features sets, we compared the 
actual results of \solver\ w.r.t. its expected performance, that is the 
ideal performance of the underlying \sunny\ algorithm. 
Test results show that the difference between simulated and actual behavior 
is minimal, thus encouraging new insights, extensions and implementations of 
this simple but effective algorithm.

\emph{Paper structure. } In Section \ref{sec:sunny} we describe the motivations and the algorithm 
underlying \sunny\ while in Section \ref{sec:validation} we explain the methodology and 
we report the results of our comparisons. In Section \ref{sec:solver} 
we introduce the CSP portfolio solver \solver\ as well as an empirical 
validation of its performance. In Section \ref{sec:related} we discuss the related 
literature while Section \ref{sec:conclusion} concludes by discussing also future work.

\section{\sunny}
\label{sec:sunny}
%In this section we will itemize five main motivations that prompted us to implement \sunny\ 
%and then we will provide a step-by-step description of the underlying algorithm.
%\subsection{Motivations}
One of the main empirical observations at the base of \sunny\
%, but also other portfolio approaches like 3S, 
is that usually combinatorial problems are extremely easy for some solvers and, at the same time, 
almost impossible to solve for others. Moreover, in case a solver is not able to solve an  
instance quickly, it is likely that such solver takes a huge amount of time to solve the instance.
A first motivation behind \sunny\ is therefore to select and schedule a subset of the 
solvers of a portfolio instead of trying to predict the ``best'' one for a given unseen 
instance. This strategy hopefully allows one to solve the same amount of 
problems minimizing the risk of choosing the wrong solver.

Another interesting consideration is that the use of 
large portfolios might not always lead to performance boost. In some cases the overabundance of solvers 
hinders the effectiveness of the considered approach. Indeed, selecting the best solver to use is 
more difficult when a big size portfolio is considered since there are more available choices. 
Despite the literature having examples of 
portfolios consisting of nearly 60 solvers \cite{NikolicMJ09}, as pointed out by 
\citeN{cpaior,pulina09} usually the best results are obtained by adopting a 
relatively small portfolio (e.g., ten or even less solvers). %: this is the second motivation underlying \sunny.

%The third 
Another motivation for \sunny\ is that --- as witnessed for instance by the good performance reached 
by \cphydra, ISAC \cite{isac}, 3S, CSHC \cite{cshc} --- the 'similarity assumption',  
stating that similar 
instances will behave similarly, is often reasonable. It thus makes sense to use algorithms 
such as $k$-NN to exploit the closeness between different instances.
% Moreover, such a kind of instance-based learning allows to classify a new instance without 
% explicitly building an off-line model: in order to compute the neighborhood of the instance 
% to classify, it is basically enough to extract (just once at the beginning) a proper set of 
% features from each of the training instances.
As a side effect, this allows to relax the off-line training phase that, as previously stated, 
makes the majority of the portfolio approaches rarely used in practice.

% Finally, one more remark concerns the extendibility and generality. Indeed, we wanted to 
% devise an algorithms portfolio that it is not confined to the resolution of CSPs but that it be 
% instead easily portable and generalizable to other domains such as for instance SAT, QBF, ASP and 
% COPs.

Starting from these assumptions and driven by these motivations we developed \sunny, 
a lazy portfolio approach for constraint solving. 

\subsection{Algorithm}
Let us explain in more detail the specifications of the algorithm. \sunny\ is the 
acronym of:
 \begin{itemize}
  \item \emph{SUb-portfolio}: for a given instance, we select a suitable 
  sub-portfolio (i.e., a subset of the constituent solvers of the portfolio)
  to run;
  \item \emph{Nearest Neighbor}: to determine the sub-portfolio we use a $k$-NN 
  algorithm that extracts from previously seen instances the $k$ instances that are 
  the closest to the instance to be solved;
  \item \emph{lazY}: no explicit prediction model is built off-line.
  %and no off-line training phase is used. 
  \end{itemize}
In a nutshell, the underlying idea behind SUNNY is therefore to minimize the 
probability of choosing the wrong solvers(s) by exploiting instance similarities 
in order to quickly get the smallest possible schedule of solvers.
%   but use the set of training examples as a case base. The only requirement is to have available 
%   for each instance of the training set both its features and the time that each solver of the 
%   portfolio takes to (possibly) solve such an instance.

The pseudo-code of \sunny\ is presented in Listing \ref{lst:sunny}.
\sunny\ takes as input the problem \texttt{inst} to be solved, the portfolio of 
solvers \texttt{solvers}, a backup solver \texttt{bkup\_solver},\footnote{A backup solver 
is a special solver of the portfolio (typically, its single best solver) aimed 
to handle exceptional circumstances (e.g.,premature failures of other solvers).} a 
parameter $\mathtt{k}$ ($\geq 1$) representing the size of the  
neighborhood to consider, a parameter $\mathtt{T}$ representing the total time 
available for running the portfolio solver, and a knowledge base $\mathtt{KB}$ of known instances for each 
of which we assume to know the features and the runtimes for every solver 
of the portfolio.

\begin{lstlisting}[mathescape, caption=\sunny\ Algorithm, label=lst:sunny, float,floatplacement=H]                  
$\sunny$($\mathtt{inst}$, $\mathtt{solvers}$, $\mathtt{bkup\_solver}$, $\mathtt{k}$, $\mathtt{T}$, $\mathtt{KB}$):
  $\mathtt{feat\_vect}$ = $\mathtt{getFeatures}$($\mathtt{inst}$, $\mathtt{KB}$)
  $\mathtt{similar\_insts}$ = $\mathtt{getNearestNeigbour}$($\mathtt{feat\_vect}$, $\mathtt{k}$, $\mathtt{KB}$)
  $\mathtt{sub\_portfolio}$ = $\mathtt{getSubPortfolio}$($\mathtt{similar\_insts}$, $\mathtt{solvers}$, $\mathtt{KB}$)
  $\mathtt{slots}$ = $\sum_{\mathtt{s} \in \mathtt{sub\_portfolio}}$ $\mathtt{getMaxSolved}$($\mathtt{s}$, $\mathtt{similar\_insts}$, $\mathtt{KB}$, $\mathtt{T}$) +
       ($\mathtt{k}$ - $\texttt{getMaxSolved}$($\mathtt{sub\_portfolio}$, $\mathtt{similar\_insts}$, $\mathtt{KB}$, $\mathtt{T}$))
  $\mathtt{time\_slot}$ = $\mathtt{T}$ / $\mathtt{slots}$
  $\mathtt{tot\_time}$ = 0
  $\mathtt{schedule}$ = {}
  $\mathtt{schedule[bkup\_solver]}$ = 0
  $\textbf{for}$ $\mathtt{solver}$ $\textbf{in}$ $\mathtt{sub\_portfolio}$:
    $\mathtt{solver\_slots}$ = $\texttt{getMaxSolved}$($\mathtt{solver}$, $\mathtt{similar\_insts}$, $\mathtt{KB}$, $\mathtt{T}$)
    $\mathtt{schedule[solver]}$ = $\mathtt{solver\_slots}$ * $\mathtt{time\_slot}$
    $\mathtt{tot\_time}$ += $\mathtt{solver\_slots}$ * $\mathtt{time\_slot}$
  $\textbf{if}$ $\mathtt{tot\_time}$ < $\mathtt{T}$:
    $\mathtt{schedule[bkup\_solver]}$ += $\mathtt{T}$ - $\mathtt{tot\_time}$
  return $\mathtt{sort}$($\mathtt{schedule}$, $\mathtt{similar\_insts}$, $\mathtt{KB}$)
\end{lstlisting}
When a new unseen instance \texttt{inst} comes, \sunny\ first extracts from it a proper set of 
features via the function $\texttt{getFeatures}$ (line 2). This function takes 
as input also the knowledge base $\mathtt{KB}$ since the extracted features need to be preprocessed 
in order to scale them in the range $[-1,1]$ and to remove the constant ones. $\texttt{getFeatures}$ returns the features vector 
\texttt{feat\_vect} of the instance \texttt{inst}. 
In line 3, function $\texttt{getNearestNeigbour}$ is used to retrieve the $\mathtt{k}$ 
nearest instances \texttt{similar\_insts} to the instance \texttt{inst} according to a certain 
distance metric (e.g., Euclidean). Then, 
in line 4 the 
function $\texttt{getSubPortfolio}$ selects the minimum subset of the portfolio that allows to 
solve the greatest number of instances in the neighborhood, by using the 
average solving time for tie-breaking.\footnote{In order to compute the average solving time, we 
assign to an instance not solved within 
the time limit $\mathtt{T}$ a solving time of $\mathtt{T}$ seconds.}

Once computed the sub-portfolio, we 
partition the time window $[0, \mathtt{T}]$ into $\mathtt{slots}$ equal time slots of size $\mathtt{T} / 
\mathtt{slots}$, where $\mathtt{slots}$ is the sum of the solved instances for each solver of the 
sub-portfolio plus the instances of \texttt{similar\_insts} that can 
not be solved within the time limit $\mathtt{T}$. In 
order to compute $\mathtt{slots}$, we use the function $\mathtt{getMaxSolved}$($\mathtt{s}$, $\mathtt{similar\_insts}$, $\mathtt{KB}$, $\mathtt{T}$) that
returns the number of instances in $\mathtt{similar\_insts}$ that a solver (or a portfolio of 
solvers) $\mathtt{s}$ is able to solve in time $\mathtt{T}$.
In lines 9--10 the associative array 
\texttt{schedule}, used to define the solvers schedules, is initialized. In 
particular, 
\texttt{schedule[s] = t} iff a time window of $t$ seconds is 
allocated to the solver $s$.

The loop enclosed between lines 11--14 assigns to each solver of the portfolio a 
number of time slots proportional to the number of instances that such 
solver can solve in the neighborhood. In lines 15--16 the remaining time slots, corresponding to the 
unsolved instances, are allocated to the backup solver.
Finally, line 17 
returns the final schedule, obtained by sorting the solvers by average solving time in 
\texttt{similar\_insts}.

\begin{example}
Let us suppose that $\mathtt{solvers} = \{s_1, s_2, s_3, s_4 \}$, $\mathtt{bkup\_solver} = 
s_3$,
$\mathtt{T} = 1800$ seconds, $\mathtt{k} = 5$, $\mathtt{similar\_insts} = \{p_1, ..., p_5\}$, and 
the run-times of 
the problems defined by $\mathtt{KB}$ as listed in Table \ref{table:runtimes}.
\begin{table}[h]
\begin{minipage}{0.45\textwidth}
\centering
  \begin{oldtabular}{|c||c|c|c|c|c||}
    \cline{2-6}
    \multicolumn{1}{c||}{} & $p_1$ & $p_2$ & $p_3$ & $p_4$ & $p_5$ \\
    \cline{1-6}\cline{1-6}
    $s_1$ & \emph{T} & \emph{T} & \textbf{3} & \emph{T} & \textbf{278} \\
    \cline{1-6}\cline{1-6}
    $s_2$ & \emph{T} &  \textbf{593} & \emph{T} & \emph{T} & \emph{T} \\
    \cline{1-6}\cline{1-6}
    $s_3$ & \emph{T} & \emph{T} &  \textbf{36} & \textbf{1452} & \emph{T} \\
    \cline{1-6}\cline{1-6}
    $s_4$ & \emph{T} & \emph{T} & \emph{T} & \textbf{122} & \textbf{60} \\
    \cline{1-6}\cline{1-6}
    \multicolumn{1}{c}{}
    \end{oldtabular}
\caption{Runtimes (in seconds).
\emph{T} indicates solver timeouts.}
\label{table:runtimes}
\end{minipage}
~\hfill~
\begin{minipage}{0.45\textwidth}
\centering
\includegraphics[height=0.1\textheight, width=0.8\textwidth,trim=1.5cm 20cm 1.3cm 0cm, 
clip]{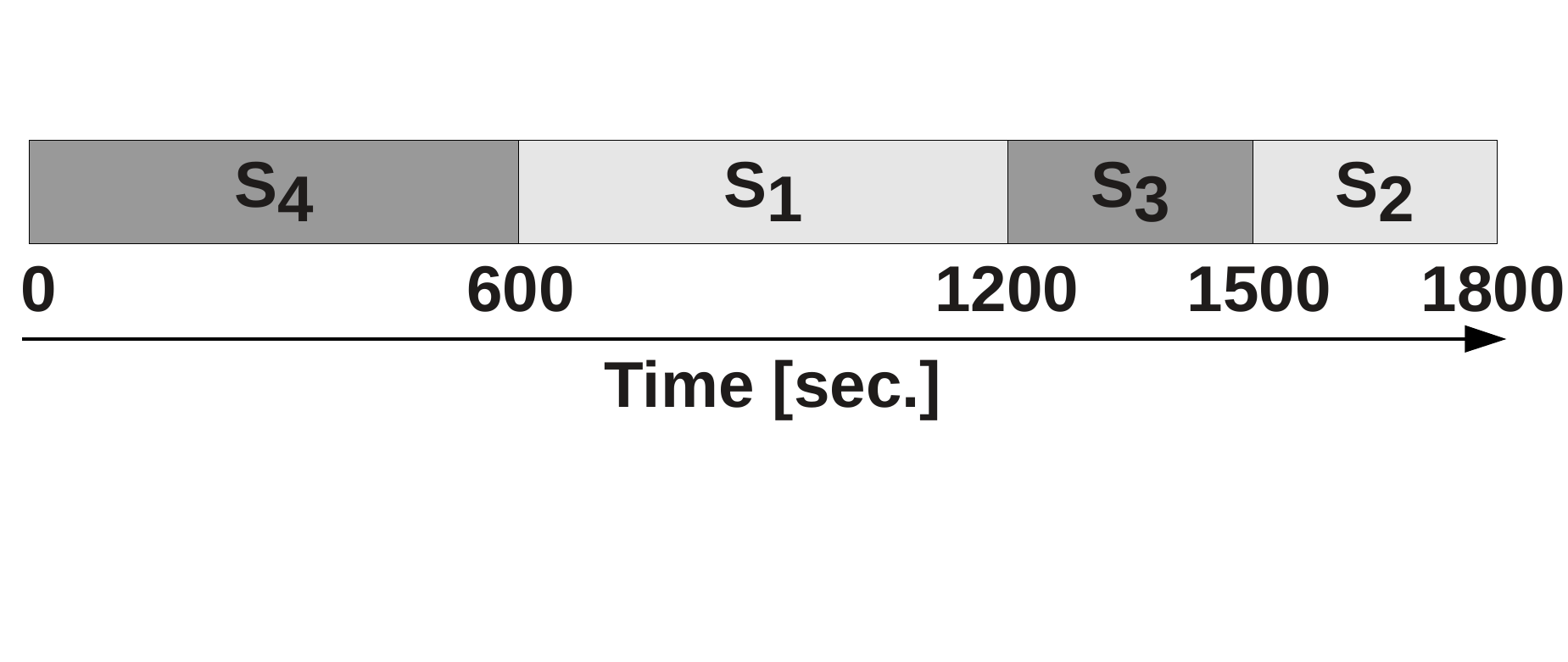}
\caption{Resulting schedule of the solvers.}
\label{fig:schedule}
\end{minipage}
\end{table}

The minimum size sub-portfolios that allow to solve the most instances 
(i.e., four instances) are $\{s_1, s_2, s_3\}$, $\{s_1, s_2, s_4\}$, and 
$\{s_2, s_3, s_4\}$. \sunny\ selects 
$\mathtt{sub\_portfolio} = \{s_1, s_2, s_4\}$ since it has a lower average solving time 
(1270.4 sec., to be precise).
% Their average solving time is respectively 1264.67, 1210.67, and 
% 1350.87 seconds. Therefore, the chosen one will be $SP_2$. 
% Now, the resulting 
% schedule will be computed proportionally to the number of instances solved.
Since $s_1$ and $s_4$ solve 2 instances, $s_2$ solves 1 instance and $p_1$ is not solved by any 
solver within $\mathtt{T}$ seconds, the time window $[0, \mathtt{T}]$ is partitioned in $2 + 2 + 1 + 1 = 6$ 
slots: 2 assigned 
to $s_1$ and $s_4$, 1 slot to $s_2$, and 1 to the backup solver $s_3$. After sorting 
the solvers by average solving time in the neighborhood we get 
the schedule illustrated in Table \ref{fig:schedule}.
\end{example}

Clearly, the proposed algorithm features a number of degrees of freedom. For instance, 
the underlying $k$-NN algorithm depends on the quality of the features extracted 
(and possibly filtered), on the choice of the $k$ parameter, and on the distance metric 
adopted.
% An interesting future direction may be consider the impact of choosing 
% different features and/or distance metrics.

A potential weakness of \sunny\ is that it could become impracticable for 
large portfolios. Indeed, in the worst case, the complexity of $\texttt{getSubPortfolio}$ 
is exponential w.r.t.~the portfolio size. However, the computation of this 
function is almost instantaneous for portfolios containing up to 15 solvers and, as later detailed, 
from an empirical point of view the sizes of the sub-portfolios are small for reasonable values of 
$k$.

Finally, let us note that the assignment of the uncovered instances of $N(p, k)$ with the 
backup solver allows the assignment of some slots to (hopefully) the most reliable 
solver. This choice obviously biases the schedule toward the backup solver, but 
experimental results have proven the effectiveness of this approach.
% 
% Note that \sunny\ has a number of degrees of freedom. For instance, is possible to 
% tune the $k$ parameter or the distance metric. Moreover, note that if $k = 1$ 
% then \sunny\ is equivalent to the simple Nearest Neighbour algorithm.
% Finally, remark that the accuracy of the $k$-NN algorithm can be severely degraded 
% by the presence of noisy or irrelevant features: is therefore fundamental to 
% extract an adequate set of features and scaling them in a reduced range.

% It is worth noticing that \sunny\ does not require any training: this allows one to 
% remove, add or 
% modify instances and solvers with a limited cost.
% The scheduling by using only the $k$-NN 
% algorithm without the need for more sophisticated and expensive techniques, like 
% for instance Mixed Integer Programming.

\section{Validation}
\label{sec:validation}
Taking as baseline the methodology and the results of \citeN{cpaior} in this section we 
present the main ingredients and the procedure that we have used for conducting 
our experiments and for evaluating the portfolio approaches, as well as 
the obtained experimental results. We would like to point out that, in order to reduce the 
computational costs,  the results 
of this Section are based on simulations. We computed the running 
times of all the solvers of the portfolio just once and used this information to  
evaluate the performance of every approach on every instance of the test 
set.
% However, as will be 
% seen in Section \ref{sec:solver}, we can assume such estimations pretty reliable.
To conduct the experiments we used Intel Dual-Core 2.93GHz computers with 3 MB 
of CPU cache, 2 GB of RAM, and Ubuntu 12.04 operating system. For keeping track 
of the solving times we considered the CPU time by exploiting the Unix 
\texttt{time} command.

\subsection{Solvers, Dataset, and Features}
\label{sec:solvers}
In order to evaluate the sensitivity of the proposed approaches w.r.t. the use of different 
solvers, we built portfolios of different size by considering all the publicly 
available and directly usable solvers of the MiniZinc Challenge 2012, namely: 
BProlog, Fzn2smt, CPX, G12/FD, G12/La\-zy\-FD, G12/MIP, Gecode, izplus, MinisatID, Mistral 
and OR-Tools. We used all of them with their default parameters, their global constraint 
redefinitions when available, and keeping track of their performances within a timeout 
of $T = 1800$ seconds.\footnote{We decided to use the same timeout of the International 
Constraint Solver Competitions.}

To conduct our experiments on a dataset of instances as realistic and 
large as possible we considered the dataset used in \citeN{featureExtractorExperiments}.
This dataset was obtained by combining 1650 instances gathered 
from the MiniZinc 1.6 benchmarks, 6 instances from the MiniZinc challenge 2012, and 6944 instances 
from the International Constraint Solver Competitions (ICSC) of 2009, discarding the ``easiest'' 
instances (i.e., those solved by Gecode in less than 2 seconds) and the ``hardest'' ones  (i.e., 
those for which the features extraction has required more than $T / 2 = 900$ seconds).
The benchmark thus obtained consisted of 4642 instances (3538 from ICSC, 6 from MiniZinc Challenge 
2012, and 1098 from MiniZinc 1.6 benchmarks).

For every instance of the benchmark we have extracted a set of 155 different 
features by exploiting the tool \tool~\cite{mzn2feat}. Among the extracted features, 144 were 
\textit{static}, i.e., obtained by parsing the source problem instance while 11 were 
\textit{dynamic}, i.e., obtained by running the Gecode solver for a short run of 2 
seconds. For further details about the benchmark and the features we refer the interested reader 
to \citeN{featureExtractorExperiments}.

Following what is usually done by the majority of current approaches, we removed all the constant 
features and we scaled their values in the range [-1, 1]. In this way
we ended up with a reduced set of 114 features.

\subsection{Portfolios Composition, Approaches, and Evaluation}
After having run every solver on each instance of the benchmark, by using a timeout of 
$T = 1800$ seconds and by keeping track of all the runtimes, we built fixed portfolios of different 
size $m = 2, \dots, 11$. More specifically, for $m = 2, \dots, 11$ the portfolio composition was 
computed by considering the portfolio of size $m$ which maximized the 
number of potential solved instances (possible ties were broken by minimizing the average solving 
time). Among all the constituent solvers, we elected MinisatID~\cite{minisatid} as a 
\textit{backup solver}, since it is the one that solved the greatest number of instances 
within the time limit $T$.

For each of such portfolios we then compared the performances of SUNNY against some of the most 
effective portfolio approaches. As done in \citeN{cpaior}, we reproduced the 
approaches of SATzilla, 3S, and CPHydra. % (see Section \ref{sec:related}). 
Note that, since 3S and SATzilla are portfolio approaches tailored 
for SAT, we had to reimplement these approaches, adapting them to our purposes.\footnote{The 
reproduction of SATzilla used in this paper differs from the original version \cite{satzilla2012} 
because ties during solvers comparison are broken by selecting the solver that in general solves 
the 
largest number of instances. Moreover no presolver selection or other parameters tuning was
performed. To reproduce 3S, instead of using the original column generation method \cite{3s}, we 
solved the scheduling problem imposing an additional constraint that requires every solver to be run 
for an integer number of seconds. No parameter tuning was performed even in this case. For more 
details we defer the interested reader to \citeN{cpaior}.}
In order to reproduce the \cphydra\ approach, we used its original algorithm without any parameter 
tuning. However, since \cphydra\ does not scale very well w.r.t.~the size of the 
portfolio,\footnote{\cphydra\ needs to solve a NP-hard 
problem to decide the schedule of solver to use. Computing the schedule can take, in few cases, 
more than half an hour for portfolios with more than 8 solvers.} we did not consider for the 
simulations the time taken to compute the schedule of the solvers to use. Thus, the 
results of \cphydra\ presented in this paper can be considered only an upper bound of its real performances.

In order to validate and test every approach on each portfolio, by following a 
common practice we used a 5-repeated 5-fold 
cross validation \cite{cross_validation}. The benchmark was randomly partitioned in 5 
disjoint folds $\Delta_1, \dots, \Delta_5$ treating in turn 
one fold $\Delta_i$, for $i = 1, \dots, 5$, as the test set and the union of the remaining folds 
$\bigcup_{j \neq i} \Delta_j$ as the training set. To avoid possible \textit{overfitting} 
problems we repeated the random generation of the folds for 5 times, thus obtaining 25 different 
training sets (consisting of about 3714 instances each) and 25 test sets 
(consisting of about 928 instances). For every instance of every test set we then computed the solving 
strategy proposed by the particular portfolio approach and we simulated it using a time cap of 
1800 seconds. For estimating the solving time we have taken into account also the time needed for 
extracting the features.
We finally evaluated the performances of every approach in terms of Ave\-ra\-ge Solving Time 
(AST) and Percentage of Solved Instances (PSI).
When a portfolio strategy was not able to solve an instance, we set its solving time to the 
time cap $T$.

\subsection{Test Results}
\begin{figure}[t]
\centering
\begin{subfigure}{0.5\textwidth}
\centering %
\includegraphics[width=\textwidth, clip, trim=1.2cm 19cm 1.5cm 0cm]{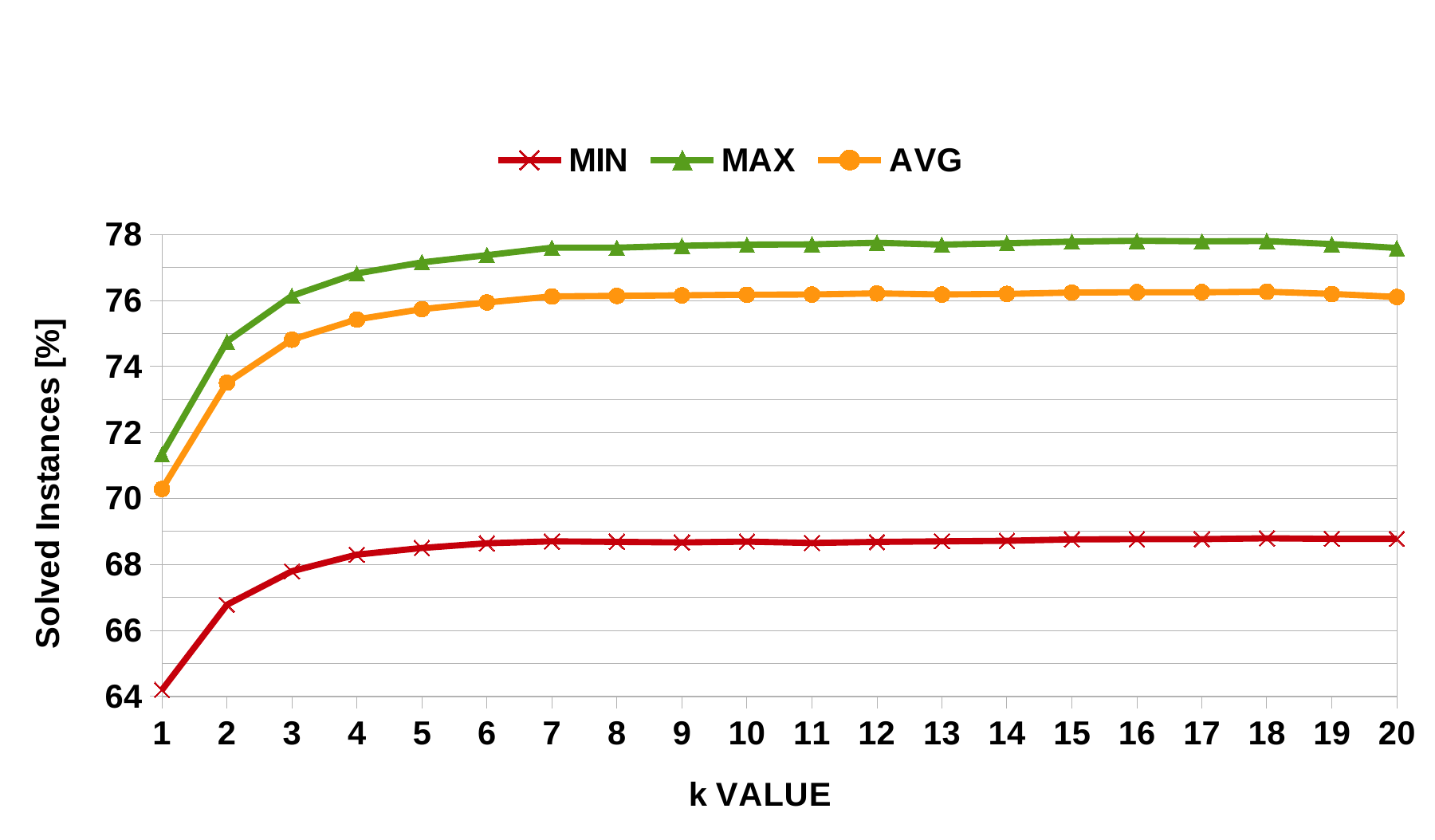}
\caption{Minimum, maximum, and average of the PSI reached by SUNNY on all the considered portfolios, 
by ranging the $k$ parameter in $[1, 20]$.}
\label{fig:k_2}
\end{subfigure}
%~\hfill~
\begin{subfigure}{0.6\textwidth}
\centering %
\includegraphics[width=\textwidth,clip, trim=1.2cm 20cm 0cm 0cm]{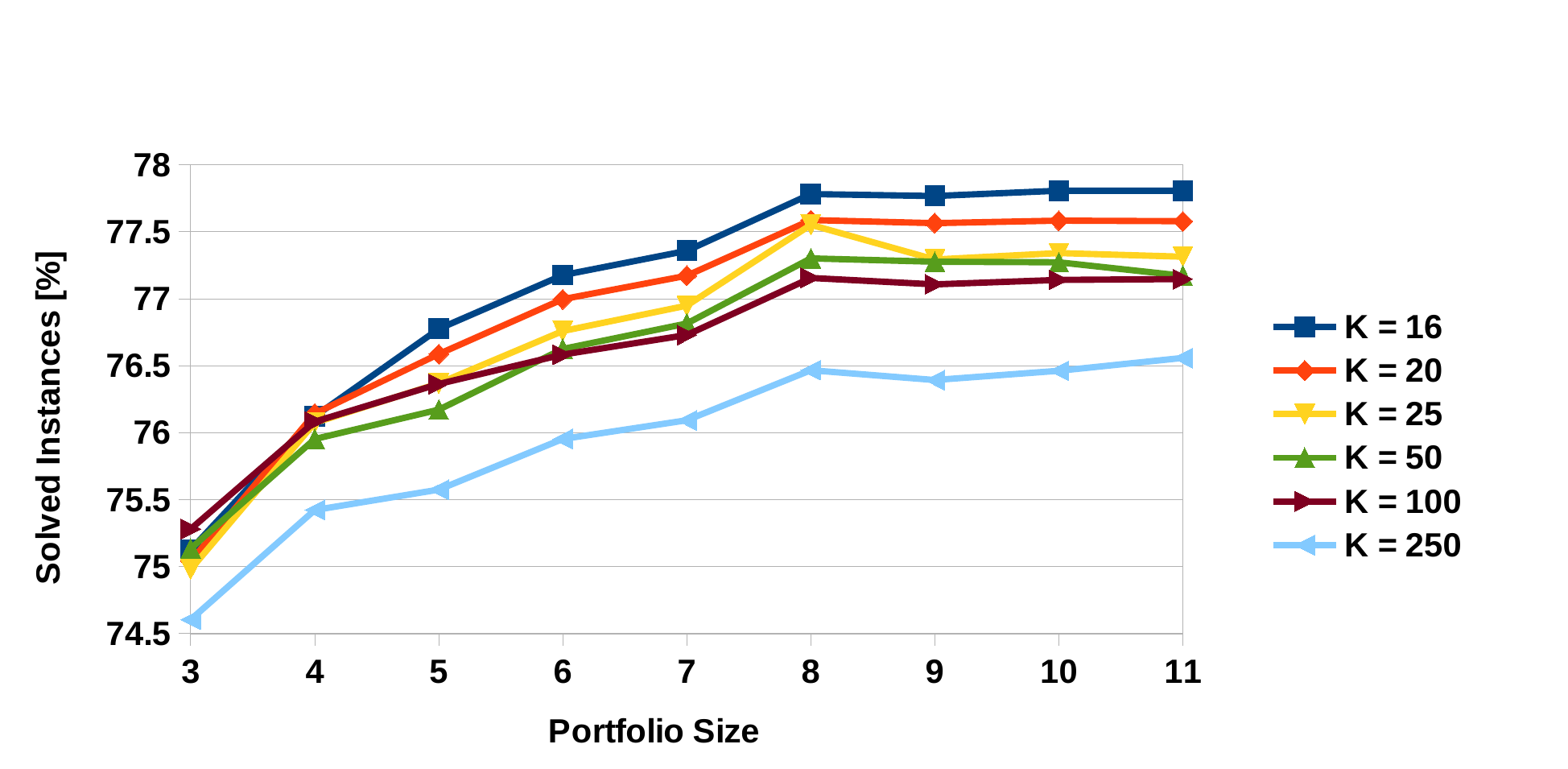}
\caption{PSI of SUNNY with $k = 16, 20, 25, 50, 100, 250$. 
%As can be observed, if $k \geq 5$ the performances are basically equivalent.
}
\label{fig:k_1}
\end{subfigure}

\caption{PSI of SUNNY varying the $k$ parameter.}
\label{fig:knn}
\end{figure}

%As pointed out at the end of Section \ref{sec:sunny}, SUNNY has some degrees of freedom. 
Before comparing \sunny\ against all the other approaches, we performed a 
sensitivity analysis by tuning the $k$ parameter.
% ~\footnote{Accordingly 
% to what is commonly done when applying the $k$-NN algorithm, we decided to use as a distance 
% metric the Euclidean distance. Note that although it is possible to build proper models 
% for predicting a good value of $k$, this task is outside the scope of our work since 
% it would mean losing the ``laziness'' of \sunny.}
As depicted in Fig. \ref{fig:k_2} the robustness of \sunny\ is reflected by the fact that 
varying the value of $k$ in $[5, 20]$ does not entail a huge 
impact in performance (i.e., less than $1\%$ of solved instances). 
The peak performance ($PSI = 77.81\%$) is reached with $k = 16$, while for $k > 20$ we 
observed a gradual performance degradation. 
For instance, by using $k = 25, 50, 100, 250, 500$ we get a maximum PSI of 
$77.59, 77.55, 77.3, 77.16, 76.56$, and $72.63$ respectively (see Figure \ref{fig:k_1}).
Of course, the robustness of SUNNY also depends on other factors: for instance, 
by using different sets of solvers or instances these assessments may be no longer be true.

\begin{figure}[t]
\centering
\begin{subfigure}{0.6\textwidth}
\centering %
\includegraphics[width=\textwidth, clip, trim=1.2cm 18.5cm 1.5cm 0.5cm]{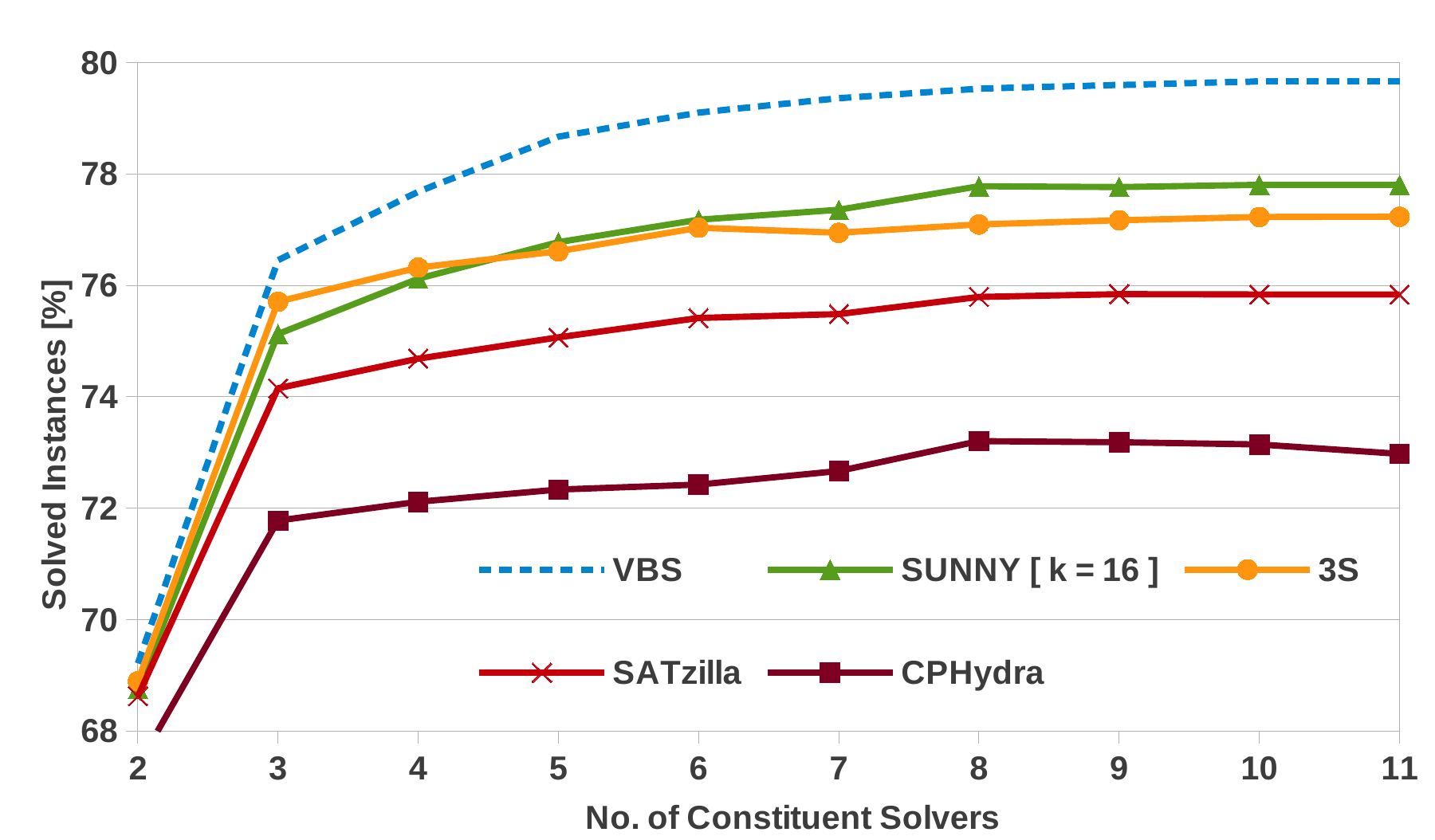}
\caption{Percentage of Solved Instances.}
\label{fig:psi}
\end{subfigure}\\
\begin{subfigure}{0.6\textwidth}
\centering %
\includegraphics[width=\textwidth, clip, trim=1.2cm 17.8cm 1.5cm 0cm]{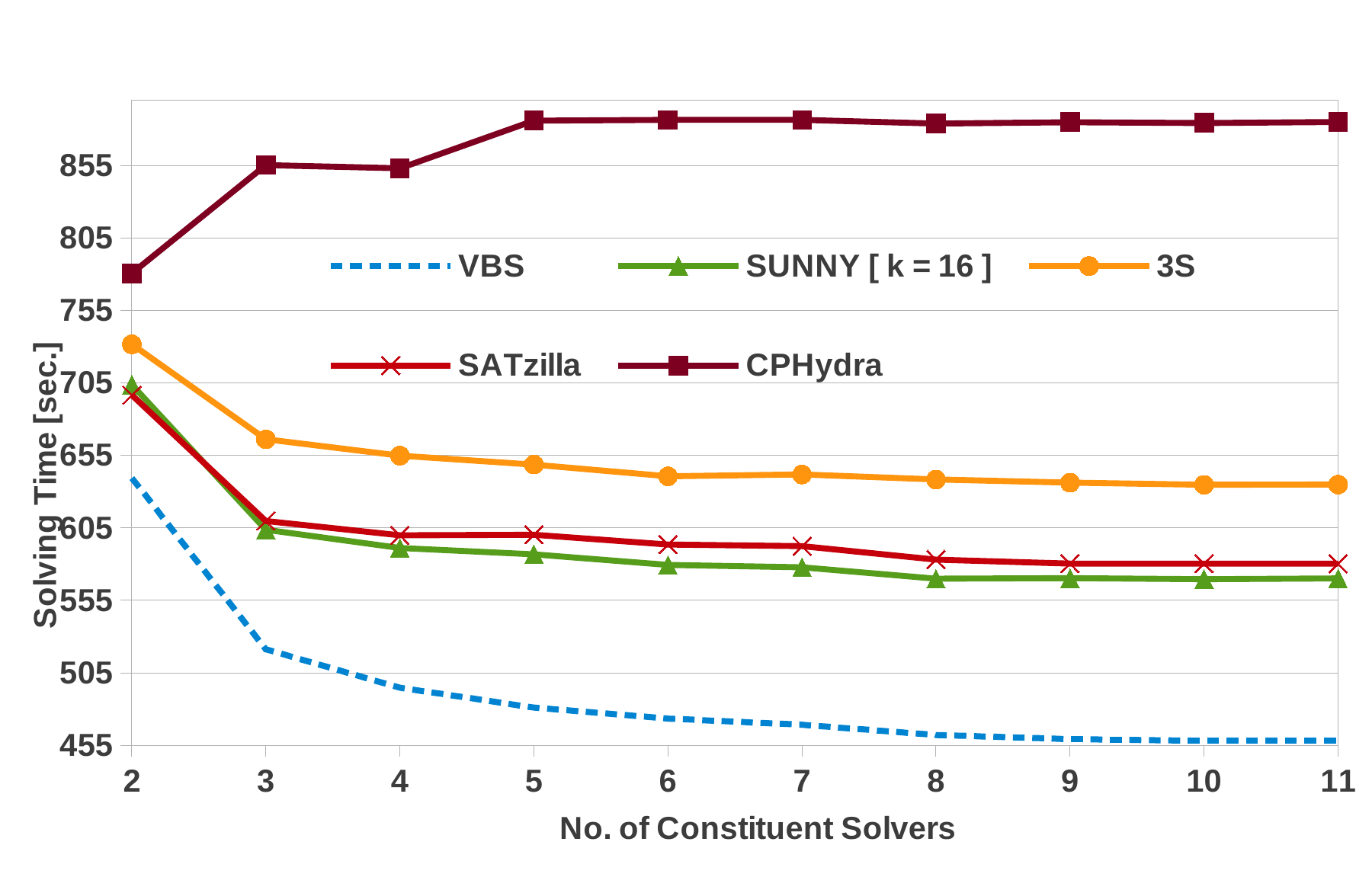}
\caption{Average Solving Time.}
\label{fig:ast}
\end{subfigure}
\caption{Performances of SUNNY, 3S, SATzilla, and CPHYdra.}
\label{fig:perf}
\end{figure}

We then selected the version of \sunny\ with $k = 16$ as the best one and we compared it against 
3S, SATzilla, and CPHydra approaches in terms of Percentage of Solved Instances (PSI) and 
Average Solving Time (AST). As a baseline, we considered the performances of the 
\textit{Virtual Best Solver} (VBS), an 'oracle' portfolio solver which 
always selects the best solver for a given instance.
As shown in Fig. \ref{fig:perf}, SUNNY turns out to be the best among all 
other approaches. Indeed, it can reach peaks of 77.81\% solved instances 
against the 76.86\% of 3S, the 75.85\% of SATzilla, and the 73.21\% of CPHYdra.  
If compared with the Single Best Solver (SBS), i.e., 
the best constituent solver in terms of number of solved instances,\footnote{The SBS for the 
selected benchmark was MinisatID~\cite{minisatid} having a PSI of 51.62\% and an AST of 950.51 seconds.} 
\sunny\ is able to close up to 
the 93.38\% of the gap between it and the VBS.
Moreover, the performances of SUNNY do not degrade at the increase of the portfolio size. Indeed 
the best performance of \sunny, both in terms of PSI and AST, is reached using a portfolio of 11 
solvers.

\begin{figure}[t]
\centering
\begin{subfigure}{0.45\textwidth}
\centering %
\includegraphics[width=\textwidth, clip, trim=1.2cm 16.5cm 1.5cm 0.5cm]{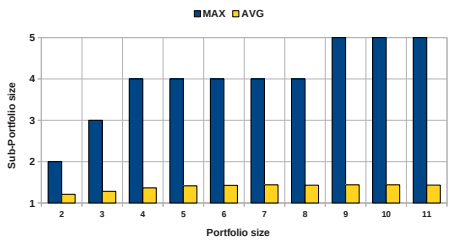}
\caption{Average and maximum sub-portfolio sizes of \sunny\ ($k = 16$).}
\label{fig:sub_size}
\end{subfigure}
~\hfill~
\begin{subfigure}{0.45\textwidth}
\centering %
\includegraphics[width=\textwidth, clip, trim=1.2cm 16.5cm 1.5cm 0.5cm]{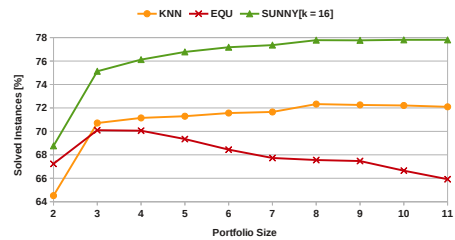}
\caption{Performances of KNN and EQU against \sunny\ ($k = 16$).}
\label{fig:knn_equ}
\end{subfigure}
\caption{\sunny\ sub-portfolio sizes and comparison against KNN and EQU.}
\end{figure}

% A possible computational issue of SUNNY concerns the computation of sub-portfolios.
% However, even if this task may be computationally intractable, we noticed that in our experiments 
% the schedules have been computed istantaneously. Moreover, 
As depicted in Fig. \ref{fig:sub_size}, 
the size of the sub-portfolios computed by \sunny\ is usually small: their average is between 
1.21 and 1.44. This means that it may be reasonable to limit the 
sub-portfolio size without compromising 
the overall performances. A further study of this issue is left as a future work.

We believe that the reasons behind the good performance of SUNNY lie in the 
quality of the features extracted as well as in the selection criteria of the 
schedules it computes. For example,
Fig. \ref{fig:knn_equ} shows the comparison of \sunny\ against two other baselines: KNN 
and EQU. KNN is the portfolio approach that for every instance selects the solver that solves more 
instances in the neighborhood, using the solving time for tie-breaking. EQU is instead an approach 
that simply allocates to every constituent solver of the portfolio an equal amount of time 
\cite{pulina09}. 
Fig. \ref{fig:knn_equ} shows that, fixing the neighborhood size to 16, the 
performances of these two approaches are significantly worse w.r.t.~to the peak performances 
of SUNNY.

\section{\solver}
\label{sec:solver}
% In Section \ref{sec:validation} we presented an empirical evaluation of \sunny\ 
% algorithm compared to the reproduction of some state of the art approach.
% We remark that, on one hand, such approaches are just ``inspired'' 
% and not the very original ones: we have reimplemented them for having baselines 
% the most accurate as possible, but it is not however the purpose of this paper 
% to state that we are far better than them. On the other, 

The results so far described are based on simulations. We decided to simulate the approaches 
because running every portfolio solver on each 
fold/repetition would take a tremendous amount of time (i.e., every single approach needs to be 
validated on $4642 * 5 = 23210$ instances).
Clearly, it is legitimate to ask if these simulations are faithful. 
Discrepancies may arise from factors not considered in the simulation like, for instance, 
the presence of memory leaks, solvers erratic behaviors, or solver faults. Moreover, since some of 
the features are dynamic, executing again the feature extraction process may lead to slightly 
different features that may cause a deviance on the expected performance.

Driven by these motivations, we therefore decided to develop and test \solver: a 
CSP portfolio built on the top of the \sunny\ algorithm. \solver\ essentially 
uses the pseudo-code of Listing \ref{lst:sunny} exploiting the features 
extractor described in \citeN{mzn2feat}. Taking as reference the results of 
Section \ref{sec:validation}, \solver\ implements SUNNY by setting $k = 16$, by exploiting  
a portfolio of 11 solvers (i.e., all the solvers listed in paragraph \ref{sec:solvers}), 
and by using MinisatID as a backup solver. 
If a scheduled solver prematurely terminates its execution,
\solver\ assigns the remaining time to the next scheduled solver if any, otherwise it 
iteratively selects the solver not yet executed that solves the greatest number of instances 
of the training set, until either the instance is solved or the timeout expires.
In order to measure and compare the actual performance of \solver\ w.r.t.~its simulated 
performance we used the same training sets, the same timeout, and the same machines adopted 
for the simulations. 

Table \ref{tab:sunny} reports a comparison between the simulated (i.e., ideal) performance of SUNNY and 
the actual performance of \solver\ for each repetition and fold in terms of PSI and AST. 
As one can see, the real performance are very close to the expected ones. 
In particular, \solver\ solves on average only 0.07\% instances less than predicted (see Table 
\ref{tab:psi}). 
There are cases in which it is also better than expected 
(see repetition 4 of Table \ref{tab:psi}).
Looking at the AST statistics (Table \ref{tab:ast}), we noted a substantial 
equivalence between the two approaches (even if on average \solver\ is slightly 
better, a mean difference of 1.78 seconds appears quite insignificant).
\begin{table}[ht]
\caption{\solver\ ideal and actual performance for each repetition and fold.}
\label{tab:sunny}
\begin{minipage}{0.45\textwidth}
\centering
\scalebox{0.55}{
\begin{oldtabular}{cc cc cc cc}
\hline
& & \multicolumn{2}{c}{\textbf{IDEAL}} & \multicolumn{2}{c}{\textbf{ACTUAL}} & \multicolumn{2}{c}{\textbf{ACTUAL $-$ IDEAL}} \\ 
 \hline
\textbf{Rep.} & \textbf{Fold} & \textbf{PSI} & \textbf{\textit{AVG}} & \textbf{PSI} & \textbf{\textit{AVG}} & \textbf{PSI} & \textbf{\textit{AVG}} \\ 
\hline
1 & 1 & 75.78 &  & 75.24 &  & $-$0.54 & \\ 
1 & 2 & 78.69 &  & 78.15 &  & $-$0.54 & \\ 
1 & 3 & 78.77 &  & 78.77 &  & 0.00 & \\ 
1 & 4 & 78.13 &  & 78.23 &  & 0.11 & \\ 
1 & 5 & 78.23 & \textit{77.92} & 78.13 & \textit{77.70} & $-$0.11 & $-$\textit{0.22}\\ 
\hline
2 & 1 & 78.90 &  & 79.22 &  & 0.32 & \\ 
2 & 2 & 76.64 &  & 76.53 &  & $-$0.11 & \\ 
2 & 3 & 78.77 &  & 78.77 &  & 0.00 & \\ 
2 & 4 & 78.56 &  & 77.80 &  & $-$0.75 & \\ 
2 & 5 & 76.08 & \textit{77.79} & 76.08 & \textit{77.68} & 0.00 & $-$\textit{0.11}\\ 
\hline
3 & 1 & 78.79 &  & 78.69 &  & $-$0.11 & \\ 
3 & 2 & 76.32 &  & 76.32 &  & 0.00 & \\ 
3 & 3 & 78.56 &  & 78.13 &  & $-$0.43 & \\ 
3 & 4 & 78.13 &  & 78.66 &  & 0.54 & \\ 
3 & 5 & 77.37 & \textit{77.83} & 77.16 & \textit{77.79} & $-$0.22 & $-$\textit{0.04}\\ 
\hline
4 & 1 & 78.04 &  & 78.04 &  & 0.00 & \\ 
4 & 2 & 75.35 &  & 75.78 &  & 0.43 & \\ 
4 & 3 & 78.66 &  & 78.56 &  & $-$0.11 & \\ 
4 & 4 & 78.23 &  & 78.23 &  & 0.00 & \\ 
4 & 5 & 78.02 & \textit{77.66} & 77.91 & \textit{77.70} & $-$0.11 & \textit{0.04}\\ 
\hline
5 & 1 & 77.93 &  & 78.04 &  & 0.11 & \\ 
5 & 2 & 78.36 &  & 78.47 &  & 0.11 & \\ 
5 & 3 & 77.05 &  & 76.94 &  & $-$0.11 & \\ 
5 & 4 & 78.56 &  & 78.45 &  & $-$0.11 & \\ 
5 & 5 & 77.26 & \textit{77.83} & 77.05 & \textit{77.79} & $-$0.22 & $-$\textit{0.04}\\ 
\hline
& &
\textbf{\textit{Total Average}} & \textbf{\textit{77.81}} & 
\textbf{\textit{Total Average}} & \textbf{\textit{77.73}} &  
\textbf{\textit{Total Average}} & \textbf{\textit{$-$0.07}}\\ 
\hline
\end{oldtabular}}
\caption{Percentage of Solved Instances.}
\label{tab:psi}
\end{minipage}
  ~\hfill~
\begin{minipage}{0.45\textwidth}
\centering
\scalebox{0.55}{
\begin{oldtabular}{cc cc cc cc}
\hline
& & \multicolumn{2}{c}{\textbf{IDEAL}} & \multicolumn{2}{c}{\textbf{ACTUAL}} & \multicolumn{2}{c}{\textbf{ACTUAL $-$ IDEAL}} \\ 
 \hline
\textbf{Rep.} & \textbf{Fold} & \textbf{AST} & \textbf{\textit{AVG}} & \textbf{AST} & \textbf{\textit{AVG}} & \textbf{AST} & \textbf{\textit{AVG}} \\ 
\hline
1 & 1 & 610.03  &                 & 618.39 &                 & 8.37    & \\ 
1 & 2 & 561.75  &                 & 567.91 &                 & 6.16    & \\ 
1 & 3 & 545.77  &                 & 549.88 &                 & 4.11    & \\ 
1 & 4 & 550.57  &                 & 545.54 &                 & $-$5.04 & \\
1 & 5 & 571.98  & \textit{568.02} & 572.29 & \textit{570.80} & 0.31    & \textit{2.78} \\ 
\hline
2 & 1 & 564.72 &  & 559.29 &  & $-$5.43\\ 
2 & 2 & 577.05 &  & 581.86 &  & 4.81\\ 
2 & 3 & 540.97 &  & 537.76 &  & $-$3.21\\ 
2 & 4 & 557.87 &  & 565.39 &  & 7.52\\ 
2 & 5 & 608.41 & \textit{569.80} & 602.52 & \textit{569.36} & $-$5.89 & \textit{-0.44} \\
\hline 
3 & 1 & 561.18 &  & 557.79 &  & $-$3.39\\ 
3 & 2 & 609.80 &  & 607.50 &  & $-$2.30\\ 
3 & 3 & 564.97 &  & 568.77 &  & 3.80\\ 
3 & 4 & 556.06 &  & 542.47 &  & $-$13.60\\ 
3 & 5 & 560.65 & \textit{570.53} & 555.87 & \textit{566.48} & $-$4.78 & \textit{-4.05} \\ 
\hline
4 & 1 & 566.71 &  & 571.09 &  & 4.38\\ 
4 & 2 & 599.00 &  & 589.31 &  & $-$9.68\\ 
4 & 3 & 551.58 &  & 549.03 &  & $-$2.55\\ 
4 & 4 & 581.12 &  & 578.40 &  & $-$2.71\\ 
4 & 5 & 552.34 & \textit{570.15} & 549.28 & \textit{567.42} & $-$3.06 & \textit{-2.73} \\ 
\hline
5 & 1 & 571.06 &  & 566.74 &  & $-$4.32\\ 
5 & 2 & 558.51 &  & 557.11 &  & $-$1.41\\ 
5 & 3 & 580.12 &  & 579.47 &  & $-$0.66\\ 
5 & 4 & 581.40 &  & 570.17 &  & $-$11.23\\ 
5 & 5 & 568.84 & \textit{571.99} & 564.09 & \textit{567.51} & $-$4.75 & \textit{-4.47} \\ 
\hline
& & \textbf{\textit{Total Average}} & \textbf{\textit{570.10}} & 
    \textbf{\textit{Total Average}} & \textbf{\textit{568.32}} & 
    \textbf{\textit{Total Average}} & \textbf{\textit{$-$1.78}}\\ 
 \hline
\end{oldtabular}}
\caption{Average Solving Time.}
\label{tab:ast}
\end{minipage}
\end{table}

In conclusion we have shown that \solver\ performance are basically equivalent 
to the expected peak performance of \sunny. In particular,
focusing on the individual performance of the constituent solvers we can argue 
that \solver\ is on average able to solve 26.11\% instances more than the Single Best Solver, 
with an average solving time of 382.59 seconds less.

The source code of \solver, as well as of 
the scripts we used to conduct the experiments, is fully available and 
downloadable at \texttt{\url{http://www.cs.unibo.it/~amadini/iclp_2014.zip}}.

\section{Related Work}
\label{sec:related}
Portfolios approaches have been used in different areas ranging from SAT to 
Constraint Optimization Problems (COPs) like Mixed Integer Programming, Scheduling, 
Most Probable Explanation (MPE) and Travel Salesman Problem (TSP).
For a survey of the different portfolio approaches studied in the literature we defer the 
interested reader to the comprehensive surveys by 
\citeN{survery_algorithm_selection,selection_survey,prediction_state_art} or more 
specifically to 
\citeN{cpaior} for CSPs, to \citeN{lion} for COPs, and to \citeN{asp_portfolio} for ASP.
Here we focus just on the most promising sequential approaches (i.e., 
winners of SAT and CSP competitions) not considering, for instance, parallel portfolio approaches 
where more than one solver is run concurrently.

\cphydra~\cite{cphydra} is 
currently the only CSP solver which uses a portfolio approach.
For the feature extraction it uses the code of Mistral, one of its constituent 
solvers, extracting from every instance 36 static and dynamic features. A $k$-nearest neighbor 
algorithm is used in order
to compute a schedule of solvers which maximizes the
chances of solving an instance within a time-out of 1800 seconds. The schedule is computed 
solving an optimization problem that is NP-hard. Nevertheless, \cphydra\ was
able to win the 2008 International CSP Solver Competition.
Unfortunately, a direct comparison between \cphydra\ and \solver\ is not immediately possible since 
\cphydra\ process only instances encoded in the (old) input format XCSP, it uses solvers 
not maintained anymore, and it does not provide an API to change the training sets used for compute 
the schedule.
% The reengineering of \cphydra\ for making the direct comparison is outside 
% the scope of this paper.
Despite the similarities between \cphydra\ and \solver, there are however some 
notable differences. Apart for the simulated performance gap in terms of PSI (\solver\ can solve 
almost 5\% instances more than the simulation of \cphydra), 
% First, the time needed for computing the schedule. When the size of the portfolio is 
% greater than 6-7 solvers, CPHydra can take a long time (even hours or days) to solve the underlying 
% MIP problem. 
% Conversely, SUNNY can almost instantaneously compute its schedule for portfolios up
% to 15 constituent solvers. 
% First, as pointed out in \cite{cpaior}, 
CPHydra does not try to minimize 
the AST since --- conversely to \sunny\ --- it doesn't use any heuristic to determine the order of the 
solvers to be run.\footnote{Conversely to the other approaches, the AST of CPHydra is not 
anti-correlated to its PSI.}
% Third, note that for 
% the simulation of 
% \cphydra\ approach we did not consider the time needed to compute the schedule of the solvers: the 
% results 
% presented in this paper can be therefore considered only an upper bound of \cphydra\ real 
% performance.

\satzilla~\cite{satzilla} is a SAT solver that relies on runtime 
prediction models to select the solver that (hopefully) has the fastest 
running time on a given problem instance.
Its last version~\cite{satzilla2012}, which consistently outperforms 
the previous ones, uses a weighted random forest approach provided with an 
explicit cost-sensitive loss function punishing misclassifications in 
direct proportion to their impact on portfolio performance. SATzilla won the 
2012 SAT Challenge in the Sequential Portfolio Track.

3S~\cite{3s} is instead a SAT solver that conjugates a 
fixed-time static solver schedule with the dynamic selection of one 
long-running component solver. 3S solves the scalability issues of \cphydra\ 
because the scheduling is computed offline and covers only 10\% 
of the time limit. If a given instance is not yet solved after the short runs, 
a designated solver is chosen at runtime (using a $k$-nearest neighbors 
algorithm) and executed. 3S originally used a portfolio of $21$ SAT solvers and 
was the best-performing dynamic portfolio at the International SAT Competition 2011.

More recently, a brand new portfolio approach was proposed by \citeN{cshc}.
This portfolio improves 3S by using its schedule for 10\% of the available time and then runs 
for the remaining time a solver selected based on a 
Cost-Sensitive 
Hierarchical Clustering (CSHC). The CSHC solver won 2 gold medals in the last SAT competition 2013.
Clearly, an evaluation of CSHC --- properly adapted to CSP --- is in our interest for its  
comparison w.r.t.~\sunny. Unfortunately, this has not been immediately possible since the code of CSHC 
was not publicly available.

% Recalling that in our reach we focus only on sequential approaches, we would 
% however mention some porfolio-based parallel SAT solvers, 
% like ManySAT~\cite{Hamadi09manysat:a}, PeneLoPe~\cite{AHJ+-12-3} and 
% ppfolio~\cite{ppfolio_web}.

% In \cite{cpaior} is reported an empirical evaluation and 
% comparison of portfolio approaches for solving CSPs. Different portfolio sizes 
% and evaluation metrics were used on a large dataset of XCSP instances taken 
% from the last two International Constraint Solver Competitions.
% In \cite{DBLP:conf/ictai/ArbelaezHS10} instead Machine Learning techniques 
% are used to enhance the performance of a single CSP solver by dynamically 
% adapting its search heuristics through Support Vector Machines.

The ASP field present a lot of similarities w.r.t.~the state of research in the CSP field. Despite 
the development of portfolios is 
not as well studied as in SAT, there exist some portfolio approaches. 
\citeN{DBLP:conf/jelia/MarateaPR12} developed ME-ASP, a multi-engine solver for propositional ASP 
programs that apply ML techniques in order to inductively choose the ``best'' 
constituent solver to be run. Analogously on what done by the configuration tool ISAC~\cite{isac} 
for SAT problems, some works try instead to exploit algorithms 
portfolio to optimally tune the parameters of ASP solvers. In 
particular, in \citeN{asp_portfolio} the solver clasp is improved by using 
Support Vector Regression for choosing a good configuration. A framework that allows to learn and 
use domain-specific heuristics for choice-point selection is instead presented in 
\citeN{journals/aicom/Balduccini11} for improving the performance of SMODEL solver. 

Related to \sunny\ is also the work presented by \citeN{DBLP:conf/iclp/HoosKSS12} which
devises an approach that takes advantage of the 
modeling and solving capacities of ASP to automatically
determine a schedule from existing benchmarking data without 
rely on any domain-specific features.
% Regarding parallel portfolios, in 
% \cite{conf/lion/YunE12} is described an approach that constructs algorithm portfolios 
% intended for parallel execution and that is based on a combination of 
% case-based reasoning, a greedy algorithm, and three heuristics.

% Finally, recent works show that the interest in algorithm runtime prediction is 
% quite general and growing. For instance, a number of tools are being developed 
% in order to improve portfolio solvers usability. \texttt{snappy} 
% (Simple Neigh\-bor\-hood-based Algorithm Portfolio in PYthon) 
% \cite{snappy} is a simple and training-less algorithm 
% portfolio which relies on a nearest neighbors prediction mechanism. 
% LLAMA (Leveraging Learning to Automatically Manage Algorithm) 
% \cite{kotthoff_llama_2013} is instead a framework that facilitates the 
% exploration of different portfolio techniques on any problem domain, by 
% supporting the most common solver selectors and possibly combining them.
% A detailed overview of the state of the art in algorithm runtime prediction is 
% provided in \cite{DBLP:journals/corr/abs-1211-0906}. For a comprehensive 
% survey on portfolio approaches applied to SAT, planning, and QBF problems we 
% refer the interested reader to the comprehensive survey \cite{selection_survey}, 
% while we mention 

\section{Conclusions and Extensions}
\label{sec:conclusion}
In this work we presented \sunny, a simple yet effective algorithm portfolio that, 
without explicitly building and learning any specific model, allows one to compute a 
sequential schedule of the constituent solvers of a portfolio for solving a given CSP.
% More precisely, \sunny\ is a lazy algorithm portfolio which exploits instances similarity to 
% guess the best solver(s) to use.
% For a given problem $p$, \sunny\ uses a $k$-NN algorithm 
% to select from a training set of known instances the subset $N(p, k)$ of 
% the $k$ instances closer to $p$. Then, it creates a schedule of solvers 
% considering the smallest sub-portfolio able to solve the maximum number of 
% instances in the neighborhood $N(p, k)$. The time allocated to each solver of the 
% sub-portfolio is proportional to the number of instances it solves in $N(p, k)$.

Despite its simplicity, preliminary results have shown that this approach is 
promising and can even outperform the state of the art CSP portfolio techniques.
To get a more realistic comparison we implemented \solver, a CSP portfolio solver 
that exploits the \sunny\ algorithm in order to solve a given MiniZinc instance by using 
a portfolio of 11 solvers. Empirical evidences confirm both the effectiveness of \sunny\ 
and the potential benefits of using a portfolio of solvers: 
indeed, \solver\ greatly outperforms each of its constituent solvers.

In view of the promising preliminary results, the possible extensions of \sunny\ are manifold.
First, its flexibility and usability may naturally lead to build effective portfolios also 
outside the CSP domain. In fact, \sunny\ may be 
easily adapted to other domains such as SAT and ASP.
As an example, \sunny\ performs well even when adapted to Constraint Optimization 
Problems~\cite{lion}.
Moreover, even if in this work we focused only on sequential portfolios, the scheduling-based 
approach of \sunny\ is well suited for the construction of parallel portfolios.
An interesting future direction may be to consider the impact of choosing 
 different features and/or distance metrics for the $k$-NN algorithm.

Constraint Logic Programming (CLP) can exploit the speed up of the search of solution provided 
by \sunny. Indeed, by adapting the same techniques described in \citeN{CLPtoGecode}, from a CLP 
problem it is 
possible to derive a MiniZinc instance that can be solved by \solver.

Another interesting direction is to dynamize \sunny\ algorithm in order to make it an online portfolio 
approach. Currently, \sunny\ is indeed based on a static and pre-computed knowledge base. A 
possible extension consists in allowing the dynamic update of the knowledge 
base, thus exploiting new incoming information such as new instances, solvers, or features.

Finally, we are planning to improve the usability and the portability of \solver, extending it to the 
resolution of also optimization problems in order to possibly enrolling it to a MiniZinc Challenge.

\bibliographystyle{acmtrans}
\bibliography{biblio}

\end{document}